\documentclass[letterpaper,10pt,conference]{ieeeconf}

\IEEEoverridecommandlockouts
\usepackage{amsmath,amssymb,amsfonts,bm,mathtools}
\usepackage{graphicx}
\usepackage{booktabs}
\usepackage{subfig}
\usepackage{cite}
\usepackage{xcolor}
\usepackage{hyperref}
\usepackage{array}

\hypersetup{
  colorlinks=true,
  linkcolor=blue,
  citecolor=orange,
  urlcolor=black
}
\newcommand{\uvec}[1]{\bm{#1}}

\newcommand{\dd}{\mathrm{d}}
\newcommand{\T}{\mathsf{T}}
\newcommand{\R}{\mathbb{R}}
\newcommand{\cL}{\mathcal{L}}
\newcommand{\cD}{\mathcal{D}}

\newcommand{\figinc}[2]{%
  \IfFileExists{#1}{\includegraphics[width=#2]{#1}}%
  {\fbox{\parbox[c][3.1cm][c]{#2}{\centering Missing figure\\\texttt{\detokenize{#1}}}}}%
}

\title{\LARGE \bf
Full Nonlinear Nonholonomic Dynamics and Motion Analysis of a 3-DoF Underactuated Spherical Rolling Robot
}
\author{
Lakshmesha Krishnapa, Ahnaf Sharaar Mazahar and Seyed Amir Tafrishi*%
\thanks{Lakshmesha Krishnapa, Ahnaf Sharaar Mazahar and Seyed Amir Tafrishi* are with the Geometric Mechanics and Mechatronics in Robotics (gm$^2$R) Lab, School of Engineering, Cardiff University, Cardiff CF24 3AA, United Kingdom. Email: \{krishnappal, mazahara, tafrishisa\}@cardiff.ac.uk}
}
\renewcommand{\baselinestretch}{.9}
\begin{document}
\maketitle
\thispagestyle{empty}
\pagestyle{empty}

\begin{abstract}
This paper presents a full nonlinear constrained dynamic model of MonoRollBot, a novel 3-DoF spherical rolling robot driven by a single motor, a lead-screw transmission, and a spring-coupled internal moving mass, together with motion analysis of its behavior. To the best of our knowledge, this is one of the first full nonlinear nonholonomic models reported for a mono-actuated, super-underactuated spherical rolling robot of this kind. Because rolling without slipping is nonholonomic, the dynamics are derived using the Lagrange--d'Alembert formulation, with the lead-screw relation imposed as a holonomic constraint and the rolling condition imposed in Pfaffian form. The formulation retains the complete generalized coordinates of shell translation, shell attitude, screw travel, nut rotation, and radial mass motion. Simulations and representative motion studies show qualitative agreement with prototype behavior and reveal how gravity, compliance, and inertia jointly shape the locomotion and motion capabilities of this strongly underactuated robot. The resulting model also provides a mechanically consistent basis for future state estimation and hybrid controller design for this nonholonomic mono-actuated rolling robot.
\end{abstract}

\section{Introduction}
Spherical rolling robots are attractive for inspection and exploration because the shell naturally protects internal hardware while enabling locomotion through rolling contact \cite{diouf2024spherical,javadi2002introducing,tafrishi2025survey}. Their actuation problem is fundamentally different from that of wheeled or legged robots: the shell is typically passive, and motion must therefore be generated indirectly through internal mechanisms. This becomes especially challenging in underactuated designs, where a small number of actuators must exploit gravity, inertia, and compliance to generate rich rolling behavior \cite{he2019underactuated,tafrishi2019design}.

Spherical rolling robots have been explored for inspection, exploration, surveillance, and operation in protected or hazardous environments because their closed shells naturally shield actuators and sensors while enabling omnidirectional rolling locomotion \cite{Armour2006,diouf2024spherical,ChasePandya2012}. Across the literature, however, the internal drive architecture varies substantially: some platforms generate motion by shifting the center of mass through pendulum-like or mass-displacement mechanisms \cite{halme1996motion,MukherjeeMinorPukrushpan2002,Bai2018}, while others rely on internal rotors, gyroscopic actuation, or wheel-based internal platforms, where motion is produced through conservation of angular momentum, reaction torques, or traction transfer inside the shell \cite{Svinin2013,KaravaevKilin2015,Morinaga2014}. These design choices strongly affect the resulting dynamics, including the degree of underactuation, the inertia coupling between shell and internal mechanism, and the way rolling constraints enter the model \cite{diouf2024spherical,hogan2015generic}. In this context, MonoRollBot is particularly interesting because it does not fit neatly into the standard pendulum-driven or rotor-driven classes: instead, it combines a single motor, a lead-screw transmission, and a spring-coupled moving mass, so that rolling emerges from a coupled redistribution of internal mass, compliant radial motion, and shell-body inertial interaction \cite{liu2025monorollbot}. This gives the robot a compact underactuated architecture with configuration-dependent inertia and gravity effects that are especially worth modeling in a full constrained nonlinear form.

From a modeling perspective, spherical rolling robots have been studied through several mechanical formulations depending on their internal actuation principle and the level of fidelity required. Early works often modeled rolling robots through reduced dynamic descriptions tailored to specific architectures, such as pendulum-driven or internally actuated mechanisms, in order to analyze locomotion and control for particular designs \cite{halme1996motion,javadi2002introducing,tafrishi2019design,tafrishi2025survey}. More broadly, recent reviews show that spherical-robot modeling remains strongly design dependent, since the internal drive concept directly affects the generalized coordinates, inertia couplings, and admissible rolling assumptions \cite{diouf2024spherical}. When rolling without slipping is treated explicitly as a velocity constraint, however, the natural framework is that of nonholonomic mechanics, where the equations are derived using constrained formulations such as the Lagrange--d'Alembert approach rather than by imposing rolling relations directly inside an unconstrained Euler--Lagrange model \cite{murray1994manipulation,bloch2003nonholonomic,hogan2015generic}. This distinction becomes particularly important for underactuated multibody spherical robots, in which internal mass redistribution, compliance, and shell motion remain strongly coupled and hardly studied. 
\begin{figure}[t!]
    \centering 
    \figinc{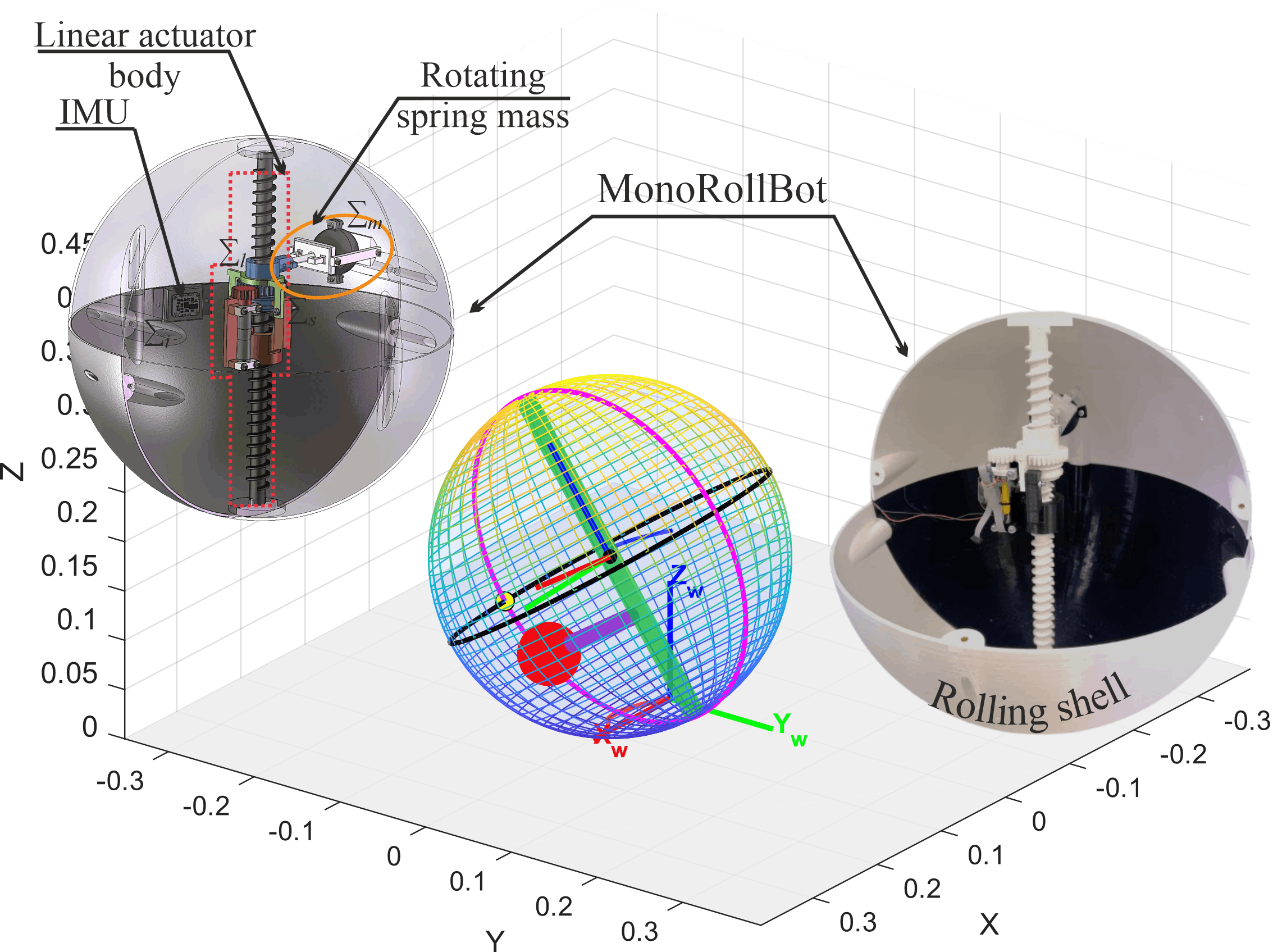}{0.42\textwidth}
    \caption{MonoRollBot robot platform and simulation model.}
    \label{fig:design}
\end{figure}

This paper develops a full nonlinear constrained model of MonoRollBot, extending our earlier design study in \cite{liu2025monorollbot}. The key technical point is that rolling without slipping is a nonholonomic velocity constraint, and the correct mechanical formulation must therefore be obtained through the constrained Lagrange--d'Alembert equations rather than from an unconstrained Euler--Lagrange model with rolling relations substituted in advance \cite{murray1994manipulation,bloch2003nonholonomic}. In contrast, the lead-screw relation between screw travel and nut rotation is holonomic, so the two constraint classes must be treated differently. Moreover, the combination of compliance, internal rotating mass, mono-actuation, and nonholonomic rolling leads to strongly coupled behavior that is not easily captured by conventional control-oriented simplifications commonly used for underactuated or nonholonomic systems \cite{Spong1994,OrioloVendittelli2005,tafrishi2025survey}. For this reason, the paper focuses not only on deriving a mechanically consistent model, but also on using it for motion-behavior analysis. The contributions of the paper are:
\begin{itemize}
    \item develop a full generalized-coordinate dynamic model of MonoRollBot using a constrained Lagrange--d'Alembert formulation with Pfaffian rolling and holonomic screw constraints;
    \item derive a compact exact mass-matrix form together with explicit conservative and damping terms;
    \item establish a simulation-ready nonlinear constrained motion model, validate it qualitatively against experiments, and use it for motion-behavior analysis.
\end{itemize}

\section{MonoRollBot Overview}
MonoRollBot is a spherical rolling robot, shown in Fig.~\ref{fig:design}, whose outer shell is completely passive and whose locomotion is generated by a single internal actuation chain. The mechanism consists of a motor-transmission unit, a lead screw, a rotating nut, and a spring-coupled moving mass arranged along the body axis. Unlike many spherical robots that use multiple internal wheels, pendula, or rotor assemblies \cite{halme1996motion,javadi2002introducing,diouf2024spherical}, MonoRollBot uses only one motor to drive the internal mechanism, making the platform mechanically compact and strongly underactuated while still capable of multi-directional rolling through internal mass redistribution \cite{liu2025monorollbot}.

Motor torque is transmitted through a gear pair to the rotating nut, while the lead screw couples nut rotation and axial translation. At the same time, the attached spring-slider mechanism allows the mass to move radially, so the internal mass evolves through coupled axial, azimuthal, and radial motion, represented here by \(d_a\), \(\theta_n\), and \(r\), respectively. As a result, MonoRollBot combines center-of-mass shift, inertial coupling, and compliance within one mechanism rather than relying on a single actuation principle \cite{tafrishi2019design,Morinaga2014}. This coupling of screw motion, nut rotation, and compliant mass displacement is what gives the robot its rolling capability and also motivates the full nonlinear constrained model developed in this paper \cite{liu2025monorollbot,bloch2003nonholonomic,murray1994manipulation}. 

\section{Nonholonomic Dynamic Modeling}
\label{sec:model_formulation}
This section develops the full nonlinear constrained dynamic model of MonoRollBot in the complete generalized coordinate space. The formulation begins by defining the reference frames, generalized coordinates, internal mass kinematics, and rolling and screw constraints, and then derives the system energetics and dissipation. The resulting equations of motion are obtained in constrained Lagrange--d'Alembert form, which is appropriate because rolling without slipping is nonholonomic while the lead-screw relation is holonomic. Finally, the model is written in an exact mass-matrix form suitable for symbolic generation and numerical solution through an augmented differential-algebraic system.
Let \(\Sigma_w\) denote the inertial world frame fixed to the ground as shown in Fig. \ref{fig:design}, with \(z_w\) normal to the flat rolling plane, and let \(\Sigma_b\) denote the shell-fixed body frame attached to the sphere center. MonoRollBot is modeled as a rigid spherical shell rolling on a flat horizontal plane without slipping. The shell is assumed isotropic in rotational inertia. The lead-screw relation is taken to be ideal, while the radial compliance is modeled by a linear spring. Dominant losses are represented by a Rayleigh dissipation function. The model targets the operating regime explored experimentally in the original MonoRollBot prototype study \cite{liu2025monorollbot}.

The moving internal mass is modeled as
$
m_c = m + m_{sb} + m_{rn},
\label{eq:mc}
$
where \(m\) is the interchangeable payload, \(m_{sb}\) is the sliding-bar mass, and \(m_{rn}\) is the rotating-nut mass. The screw mass \(m_{sl}\) is not translated by the lead and is therefore absorbed into the reflected inertial effects. To represent the finite internal travel, we define an effective core-clearance radius \(R_c\) and the offset
$
z_{\mathrm{off}} = R - R_c.
\label{eq:zoff}
$
With this convention, \(d_a=0\) corresponds to the lower internal travel limit and \(d_a=2z_{\mathrm{off}}\) corresponds to the upper limit.
\begin{table}[t]
\centering
\caption{MonoRollBot parameters and design details.}
\label{tab:params}
\footnotesize
\begin{tabular}{lll}
\toprule
Symbol & Value & Meaning \\
\midrule
$R$ & $0.17$ m & Sphere radius \\
$R_c$ & $0.03$ m & Effective core radius \\
$z_{\mathrm{off}}$ & $0.14$ m & Axial offset $R-R_c$ \\
$m_s$ & $1.00$ kg & Shell mass \\
$m_{sb}$ & $0.028$ kg & Sliding-bar mass \\
$m_{rn}$ & $0.020$ kg & Rotating-nut mass \\
$m_{sl}$ & $0.085$ kg & Screw mass \\
$l$ & $20$ mm & Screw lead \\
$P$ & $10$ mm & Screw pitch \\
$\eta_m$ & $17$ & Motor gear teeth \\
$\eta_n$ & $34$ & Nut gear teeth \\
$N_{gb}$ & $62{:}1$ & Motor reduction ratio \\
$g_{mn}$ & $\eta_m/\eta_n = 1/2$ & Gear-pair ratio \\
$a$ & $l/(2\pi)=3.1831\times10^{-3}$ m/rad & Screw factor \\
$V_{nom}$ & $6$ V & Motor nominal voltage \\
$I_{nom}$ & $0.3$ A & Motor nominal current \\
$m$ & $20,35,50,70$ g & Payload sweep \\
$k_s$ & $160,200,300$ N/m & Stiffness sweep \\
\bottomrule
\end{tabular}
\end{table}
Also, the full generalized coordinate vector is
\begin{equation}
\uvec{q}=
\begin{bmatrix}
x & y & \alpha & \beta & \gamma & d_a & \theta_n & r
\end{bmatrix}^{\T},
\label{eq:qfull}
\end{equation}
where \((x,y)\) are the planar coordinates of the sphere center in \(\Sigma_w\), \((\alpha,\beta,\gamma)\) are the Z--Y--X Euler angles of the shell, \(d_a\) is the axial screw travel measured from the lower end, \(\theta_n\) is the rotating-nut angle, and \(r\) is the radial displacement of the internal mass.

To derive the kinematis of rotating sphere, the shell attitude with respect to \(\Sigma_w\) is parameterized by
\begin{equation}
\uvec{R}_{wb}(\alpha,\beta,\gamma)=\uvec{R}_z(\gamma)\uvec{R}_y(\beta)\uvec{R}_x(\alpha),
\label{eq:RIB}
\end{equation}
and the shell angular velocity expressed in \(\Sigma_b\) is
\begin{equation}
\uvec{\omega}_B=
\uvec{E}(\alpha,\beta)
\begin{bmatrix}
\dot\alpha\\ \dot\beta\\ \dot\gamma
\end{bmatrix},
\uvec{E}(\alpha,\beta)=
\begin{bmatrix}
1 & 0 & -\sin\beta \\
0 & \cos\alpha & \sin\alpha\cos\beta \\
0 & -\sin\alpha & \cos\alpha\cos\beta
\end{bmatrix}
\label{eq:omega}
\end{equation}
Now to obtain kinematic of rotating mass, first, let the sphere-center position in \(\Sigma_w\) be
\begin{equation}
\uvec{r}_O=
\begin{bmatrix}
x & y & 0
\end{bmatrix}^{\T},
\label{eq:rO}
\end{equation}
where the constant shell-center height \(R\) along \(z_w\) is omitted from the potential since it does not affect the equations of motion. Then, the internal moving-mass position in \(\Sigma_b\) is
\begin{equation}
\uvec{p}_c^B(\uvec{q})=
\begin{bmatrix}
r\cos\theta_n\\
r\sin\theta_n\\
d_a-z_{\mathrm{off}}
\end{bmatrix},
\label{eq:pcb}
\end{equation}
where \(d_a-z_{\mathrm{off}}\) is the centered body-frame axial coordinate. The position of the internal mass with respect to inertia is 
\begin{equation}
\uvec{P}_c(\uvec{q})=\uvec{r}_O+\uvec{R}_{wb}(\uvec{q})\,\uvec{p}_c^B(\uvec{q}),
\label{eq:Pc}
\end{equation}
with inertial velocity of the core 
\begin{equation}
\dot{\uvec{P}}_c=
\dot{\uvec{r}}_O+\uvec{R}_{wb}\left(\dot{\uvec{p}}_c^B+\uvec{\omega}_B\times\uvec{p}_c^B\right).
\label{eq:Pcdot}
\end{equation}

For nonholonomic constraint due to rolling on a flat plane, the no-slip conditions are obtained
\begin{align}
&\dot x - R\left(\dot\beta\cos\alpha+\dot\gamma\sin\alpha\sin\beta\right)=0,\nonumber\\
&\dot y - R\left(-\dot\beta\sin\alpha+\dot\gamma\cos\alpha\sin\beta\right)=0.
\label{eq:nsy}
\end{align}
These are written compactly as the Pfaffian rolling constraint \cite{Camicia2000}
\begin{equation}
\uvec{A}(\uvec{q})\dot{\uvec{q}}=\uvec{0},
\label{eq:Aqd}
\end{equation}
with
\begin{equation}
\uvec{A}(\uvec{q})=
\begin{bmatrix}
1 & 0 & 0 & -R\cos\alpha & -R\sin\alpha\sin\beta & 0 & 0 & 0\\
0 & 1 & 0 &  R\sin\alpha & -R\cos\alpha\sin\beta & 0 & 0 & 0
\end{bmatrix}.
\label{eq:Amat}
\end{equation}
Also, note that the Euler angles in \eqref{eq:RIB} parameterize shell attitude, but they do not enforce rolling. For more general spherical-robot motion on curved surfaces, contact kinematics can be treated more broadly \cite{hogan2015generic}; for the present flat-ground problem, \eqref{eq:Amat} is sufficient.

To dervie motion equation, first, the kinetic energy of the full system is based on linear and angular momentum 
\begin{equation}
T=
\frac{1}{2}m_s\|\dot{\uvec{r}}_O\|^2
+\frac{1}{2}I_s\|\uvec{\omega}_B\|^2
+\frac{1}{2}m_c\|\dot{\uvec{P}}_c\|^2
+\frac{1}{2}I_c\dot\theta_n^2,
\label{eq:Tfull}
\end{equation}
and the potential energy is
\begin{equation}
V=
\frac{1}{2}k_s r^2 + m_c g z_c,
\label{eq:Vfull}
\end{equation}
where
\begin{equation}
z_c =
(d_a-z_{\mathrm{off}})\cos\alpha\cos\beta
-r\sin\beta\cos\theta_n
+r\cos\beta\sin\alpha\sin\theta_n.
\label{eq:zc}
\end{equation}
The Lagrangian function based on (\ref{eq:Aqd}), (\ref{eq:Tfull}) and (\ref{eq:Vfull}) is therefore
\begin{equation}
\cL(\uvec{q},\dot{\uvec{q}})=T(\uvec{q},\dot{\uvec{q}})-V(\uvec{q}).
\label{eq:Lagrangian}
\end{equation}
We also have friction effect that we modeled by the Rayleigh dissipation function
\begin{equation}
\cD=
\frac{1}{2}c_s(\dot\alpha^2+\dot\beta^2+\dot\gamma^2)
+\frac{1}{2}c_d\dot d_a^2
+\frac{1}{2}c_\theta\dot\theta_n^2
+\frac{1}{2}c_r\dot r^2.
\label{eq:Rayleigh}
\end{equation}

Because rolling without slipping is nonholonomic, the equations of motion are written with inspiration from constrained Lagrange--d'Alembert  \cite{murray1994manipulation,bloch2003nonholonomic} using (\ref{eq:Lagrangian}) and (\ref{eq:Rayleigh}) as
\begin{equation}
\frac{\dd}{\dd t}\left(\frac{\partial \cL}{\partial \dot{\uvec{q}}}\right)
-
\frac{\partial \cL}{\partial \uvec{q}}
+
\frac{\partial \cD}{\partial \dot{\uvec{q}}}
=
\uvec{Q} + \uvec{A}(\uvec{q})^\T \bm{\lambda} + \uvec{J}_h(\uvec{q})^\T \mu,
\label{eq:lda_full}
\end{equation}
subject to
\begin{equation}
\uvec{A}(\uvec{q})\dot{\uvec{q}}=\uvec{0},
\qquad
\phi_h(\uvec{q})=0,
\label{eq:constraints_full}
\end{equation}
where \(\bm{\lambda}\in\R^2\) is the nonholonomic multiplier vector and \(\mu\in\R\) is the holonomic multiplier associated with the screw relation. Then, the full constrained dynamics is obtained in following final form
\begin{align}
\begin{bmatrix}
\uvec{M}(\uvec{q}) & -\uvec{A}(\uvec{q})^\T & -\uvec{J}_h^\T\\
\uvec{A}(\uvec{q}) & \uvec{0} & \uvec{0}\\
\uvec{J}_h & \uvec{0} & 0
\end{bmatrix}
\begin{bmatrix}
\ddot{\uvec{q}}\\
\bm{\lambda}\\
\mu
\end{bmatrix}
\nonumber\\=
\begin{bmatrix}
\uvec{Q}(\tau_m)-\uvec{h}_s(\uvec{q},\dot{\uvec{q}})-\uvec{d}_q(\dot{\uvec{q}})\\
-\dot{\uvec{A}}(\uvec{q},\dot{\uvec{q}})\dot{\uvec{q}}\\
0
\end{bmatrix},
\label{eq:augmented_system}
\end{align}
where \(\uvec{M}(\uvec{q})\) is the full inertia matrix, \(\uvec{d}_q(\dot{\uvec{q}})=\partial\mathcal{D}/\partial\dot{\uvec{q}}\) is the generalized damping vector, \(\bm{\lambda}\in\mathbb{R}^2\) and \(\mu\in\mathbb{R}\) are the nonholonomic and holonomic multipliers, and
\begin{equation}
\uvec{Q}(\tau_m)=
\begin{bmatrix}
0&0&0&0&0&0&\tau_n&0
\end{bmatrix}^{\T},
\tau_n=\eta_g g_r \tau_m ,
\label{eq:Qfull}
\end{equation}
so the motor torque acts on the nut coordinate \(\theta_n\), while the axial screw reaction is enforced through the holonomic constraint. The screw Jacobian is
\begin{equation}
\uvec{J}_h(\uvec{q})=
\frac{\partial \phi_h}{\partial \uvec{q}}=
\begin{bmatrix}
0 & 0 & 0 & 0 & 0 & 1 & -a & 0
\end{bmatrix},
\label{eq:Jhol_compact}
\end{equation}
corresponding to \(d_a-a\theta_n=0\) and $a=l/2\pi$. Eq.~\eqref{eq:augmented_system} therefore enforces the rolling Pfaffian constraint and the screw relation directly at the acceleration level in one linear solve.
where inertia matrix terms are
\begingroup
\footnotesize
\begin{subequations}
\label{eq:M_entries_full_compact}
\begin{align}
&M_{11}=m_c+m_s,\;
M_{12}=0,\;
M_{13}=m_c\!\left(\Delta\Psi_9+r s_\theta\Psi_{10}\right), \nonumber\\
&M_{14}=m_c c_\gamma\Psi_1,\;
M_{15}=m_c\!\left(\Delta\Psi_3-r\Psi_5\right),\;
M_{16}=m_c\Psi_2, \nonumber\\
&M_{17}=-m_c r\Psi_6,\;
M_{18}=m_c\Psi_4,\;
M_{22}=m_c+m_s, \nonumber\\[0.4mm]
&M_{23}=m_c\!\left(-\Delta\Psi_{11}+r s_\theta\Psi_8\right),\;
M_{24}=m_c s_\gamma\Psi_1,\; \nonumber\\
&M_{25}=m_c\!\left(\Delta\Psi_2+r\Psi_4\right),M_{26}=m_c\Psi_8,
M_{27}=m_c r\Psi_7,
M_{28}=m_c\Psi_5, \nonumber\\[0.4mm]
&M_{33}=I_s+m_c\!\left(\Delta^2+r^2 s_\theta^2\right),\;
M_{34}=m_c r c_\theta\!\left(\Delta s_\alpha-r c_\alpha s_\theta\right), \nonumber\\
&M_{35}=-I_s s_\beta
-m_c\!\left(s_\beta\Delta^2+\Delta r c_\alpha c_\beta c_\theta+r^2 s_\beta s_\theta^2
+r^2 c_\beta s_\alpha s_\theta c_\theta\right), \nonumber\\
&M_{36}=m_c r s_\theta,\;
M_{37}=-m_c r c_\theta\Delta,\;
M_{38}=-m_c s_\theta\Delta, \nonumber\\[0.4mm]
&M_{44}=I_s+m_c\!\left(\Delta^2 c_\alpha^2+r^2\!\left(1-c_\alpha^2 s_\theta^2\right)
+2\Delta r c_\alpha s_\alpha s_\theta\right), \nonumber\\
&M_{45}=\tfrac{1}{2}m_c\Delta^2\sin(2\alpha)c_\beta
-\tfrac{1}{2}m_c r^2\sin(2\alpha)c_\beta s_\theta^2\nonumber\\
&+\tfrac{1}{2}m_c r^2\sin(2\theta_n)c_\alpha s_\beta 
-m_c\Delta r s_\alpha s_\beta c_\theta
-m_c\Delta r\cos(2\alpha)c_\beta s_\theta, \nonumber\\
&M_{46}=-m_c r c_\alpha c_\theta,\;
M_{47}=-m_c r\!\left(r s_\alpha+\Delta c_\alpha s_\theta\right),\;
M_{48}=m_c\Delta c_\alpha c_\theta, \nonumber\\[0.4mm]
&M_{55}=I_s+m_c\!\Bigl[\Delta^2\!\left(1-c_\alpha^2 c_\beta^2\right)
+2\Delta r c_\alpha c_\beta\Psi_{13}
+r^2\!\left(c_\alpha^2 c_\beta^2+\Psi_{12}^2\right)\Bigr], \nonumber\\
&M_{56}=-m_c r\Psi_{12},\;
M_{57}=m_c r\!\left(r c_\alpha c_\beta+\Delta\Psi_{13}\right),\;
M_{58}=m_c\Delta\Psi_{12}, \nonumber\\[0.4mm]
&M_{66}=m_c,\;
M_{67}=0,\;
M_{68}=0,\;
M_{77}=m_c r^2+I_c,\;\nonumber\\
&M_{78}=0,\;
M_{88}=m_c.
\end{align}
\end{subequations}
\endgroup
Here, we have $\Delta:=d_a-z_{\mathrm{off}},$ with the shorthand notation
\begin{align}
&c_\alpha=\cos\alpha,\; s_\alpha=\sin\alpha,\;
c_\beta=\cos\beta,\; s_\beta=\sin\beta, \nonumber\\
&c_\gamma=\cos\gamma,\; s_\gamma=\sin\gamma,\;
c_\theta=\cos\theta_n,\; s_\theta=\sin\theta_n,
\label{eq:trig_shorthand}
\end{align}
and the auxiliary terms
\begingroup
\small
\begin{align}
&\Psi_1=\Delta c_\alpha c_\beta-r s_\beta c_\theta+r c_\beta s_\alpha s_\theta,\;
\Psi_2=s_\alpha s_\gamma+c_\alpha c_\gamma s_\beta, \nonumber\\
&\Psi_3=s_\alpha c_\gamma-c_\alpha s_\beta s_\gamma,\;
\Psi_4=c_\beta c_\gamma c_\theta-c_\alpha s_\gamma s_\theta+c_\gamma s_\alpha s_\beta s_\theta, \nonumber\\
&\Psi_5=c_\alpha c_\gamma s_\theta+c_\beta c_\theta s_\gamma+s_\alpha s_\beta s_\gamma s_\theta, \nonumber\\
&\Psi_6=c_\alpha c_\theta s_\gamma+c_\beta c_\gamma s_\theta-c_\gamma s_\alpha s_\beta c_\theta, \nonumber\\
&\Psi_7=c_\alpha c_\gamma c_\theta-c_\beta s_\gamma s_\theta+s_\alpha s_\beta c_\theta s_\gamma,\;
\Psi_8=c_\alpha s_\beta s_\gamma-c_\gamma s_\alpha, \nonumber\\
&\Psi_9=c_\alpha s_\gamma-s_\alpha s_\beta c_\gamma,
\Psi_{10}=s_\alpha s_\gamma+c_\alpha s_\beta c_\gamma,
\Psi_{11}=c_\alpha c_\gamma \nonumber\\
&+s_\alpha s_\beta s_\gamma,\Psi_{12}=s_\beta s_\theta+c_\beta s_\alpha c_\theta,\;
\Psi_{13}=s_\beta c_\theta-c_\beta s_\alpha s_\theta .
\label{eq:Psi13}
\end{align}
\endgroup
\begin{figure*}[t!]
    \centering
     \includegraphics[width=2.0in]{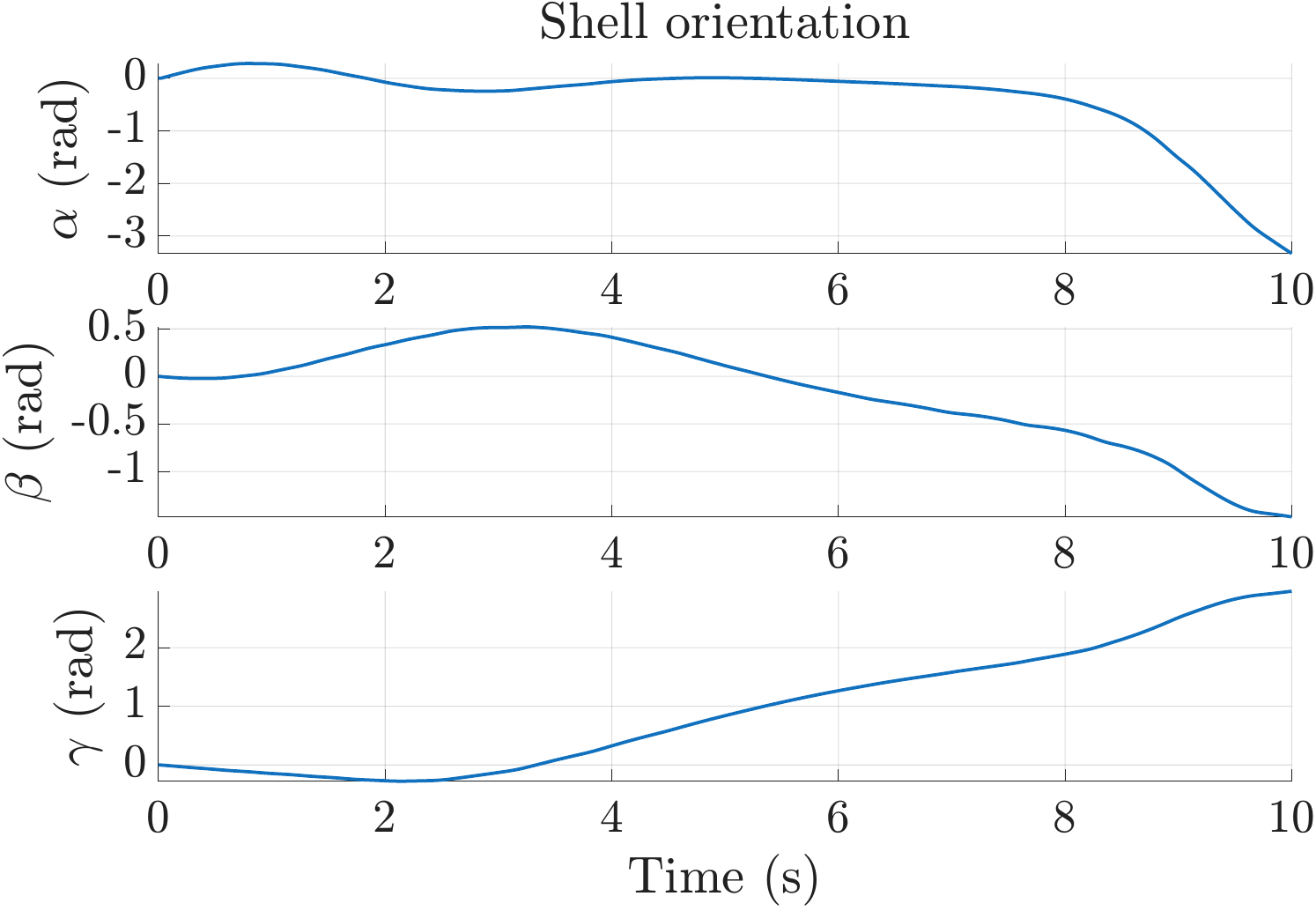}
    \includegraphics[width=2.0in]{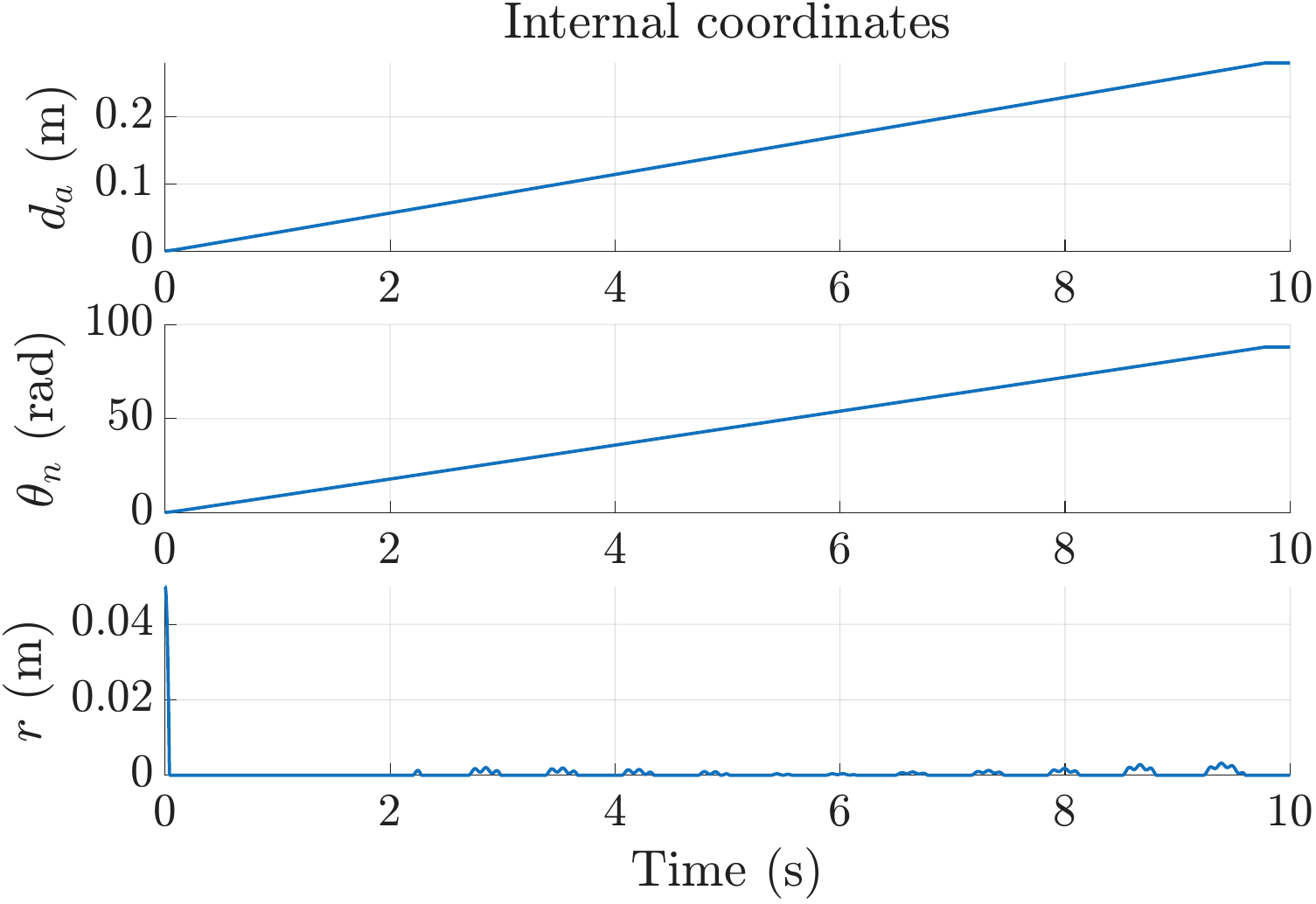}
    \includegraphics[width=2.0in]{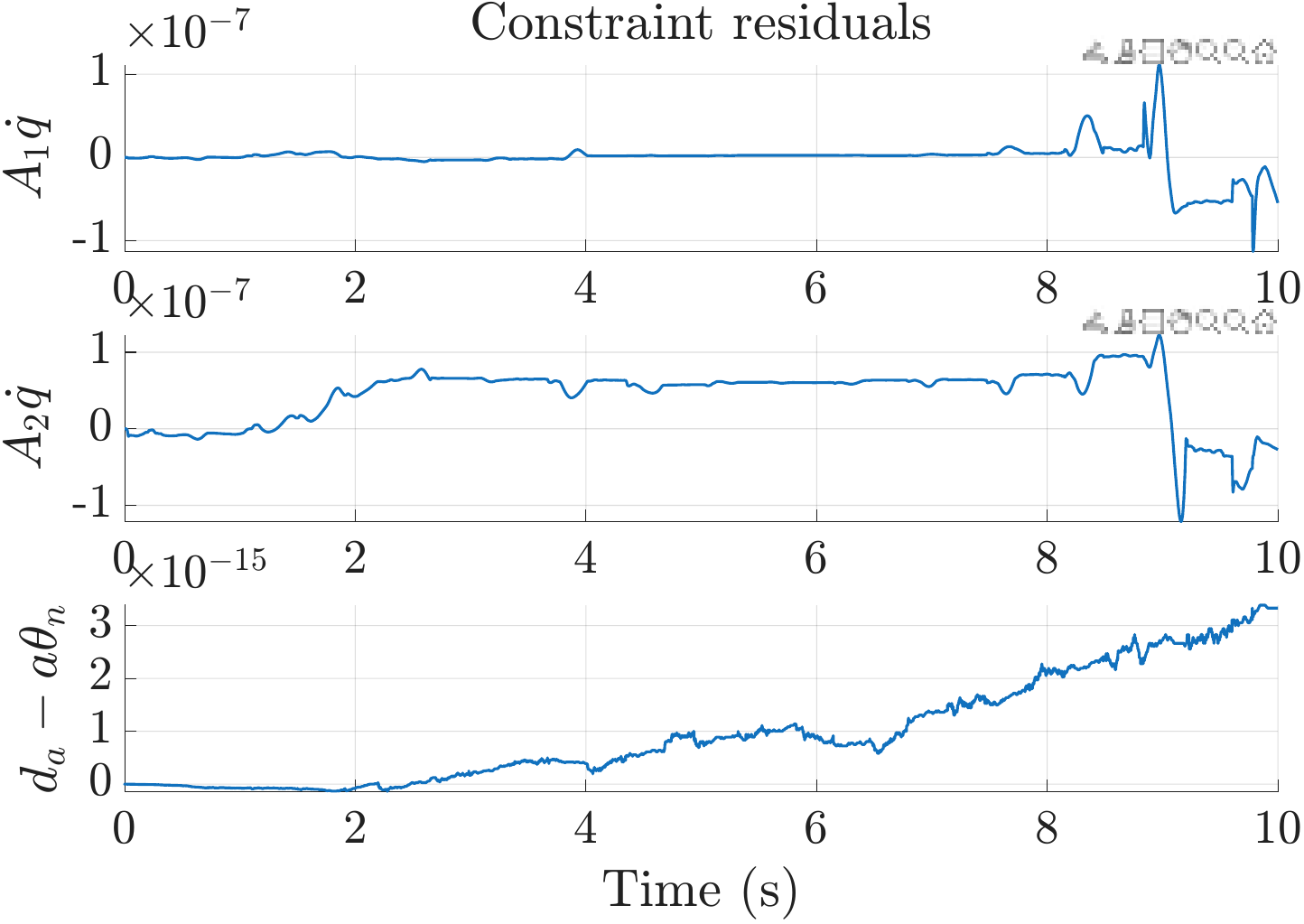}
    \caption{Representative validation-case simulation. From right to left: internal coordinates and constraint residuals, planar trajectory, and shell orientation. Small residuals confirm accurate enforcement of the rolling and screw constraints.
}
    \label{fig:fullstate_validation}
\end{figure*}
Also, the nonlinear vector \(\uvec{h}_s(\uvec{q},\dot{\uvec{q}})\) is decomposed as
\begin{equation}
\uvec{h}_s(\uvec{q},\dot{\uvec{q}})
=
\uvec{c}_q(\uvec{q},\dot{\uvec{q}})
+
\uvec{g}_q(\uvec{q}),
\label{eq:hsplit}
\end{equation}
where \(\uvec{c}_q\) contains the quadratic-velocity terms and \(\uvec{g}_q\) collects the conservative configuration terms obtained from the potential energy. The quadratic-velocity vector is written in Christoffel form as
\begin{equation}
\big[\uvec{c}_q(\uvec{q},\dot{\uvec{q}})\big]_i
=
\sum_{j=1}^{8}\sum_{k=1}^{8}
\Gamma_{ijk}(\uvec{q})\,\dot{q}_j\dot{q}_k,
\qquad i=1,\dots,8,
\label{eq:cq_component}
\end{equation}
with Christoffel symbols of the first kind
\begin{equation}
\Gamma_{ijk}(\uvec{q})
=
\frac{1}{2}
\left(
\frac{\partial M_{ij}}{\partial q_k}
+
\frac{\partial M_{ik}}{\partial q_j}
-
\frac{\partial M_{jk}}{\partial q_i}
\right).
\label{eq:christoffel_first}
\end{equation}

\begin{figure}[t!]
    \centering
    \includegraphics[width=3.1 in,height=5.5in]{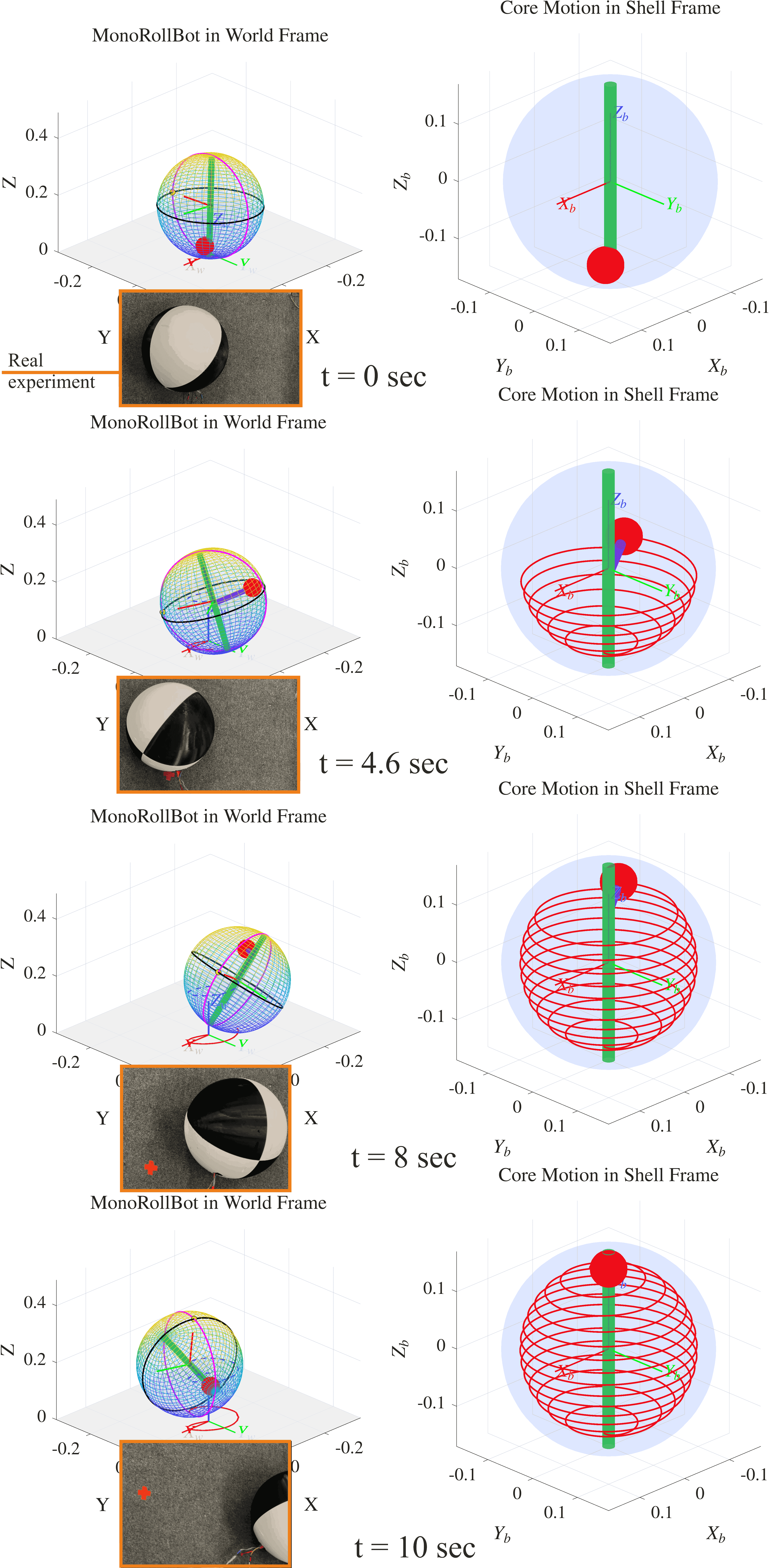}
  \caption{Representative simulated motion snapshots (with real robot corresponding experimented result) at selected times. Left: shell motion and rolling trajectory in the world frame \(\Sigma_w\). Right: internal mass motion in the shell-fixed frame \(\Sigma_b\).}
    \label{fig:sim_snapshots}
\end{figure}
\begin{figure}[t!]
    \centering
    \includegraphics[width=2.5  in]{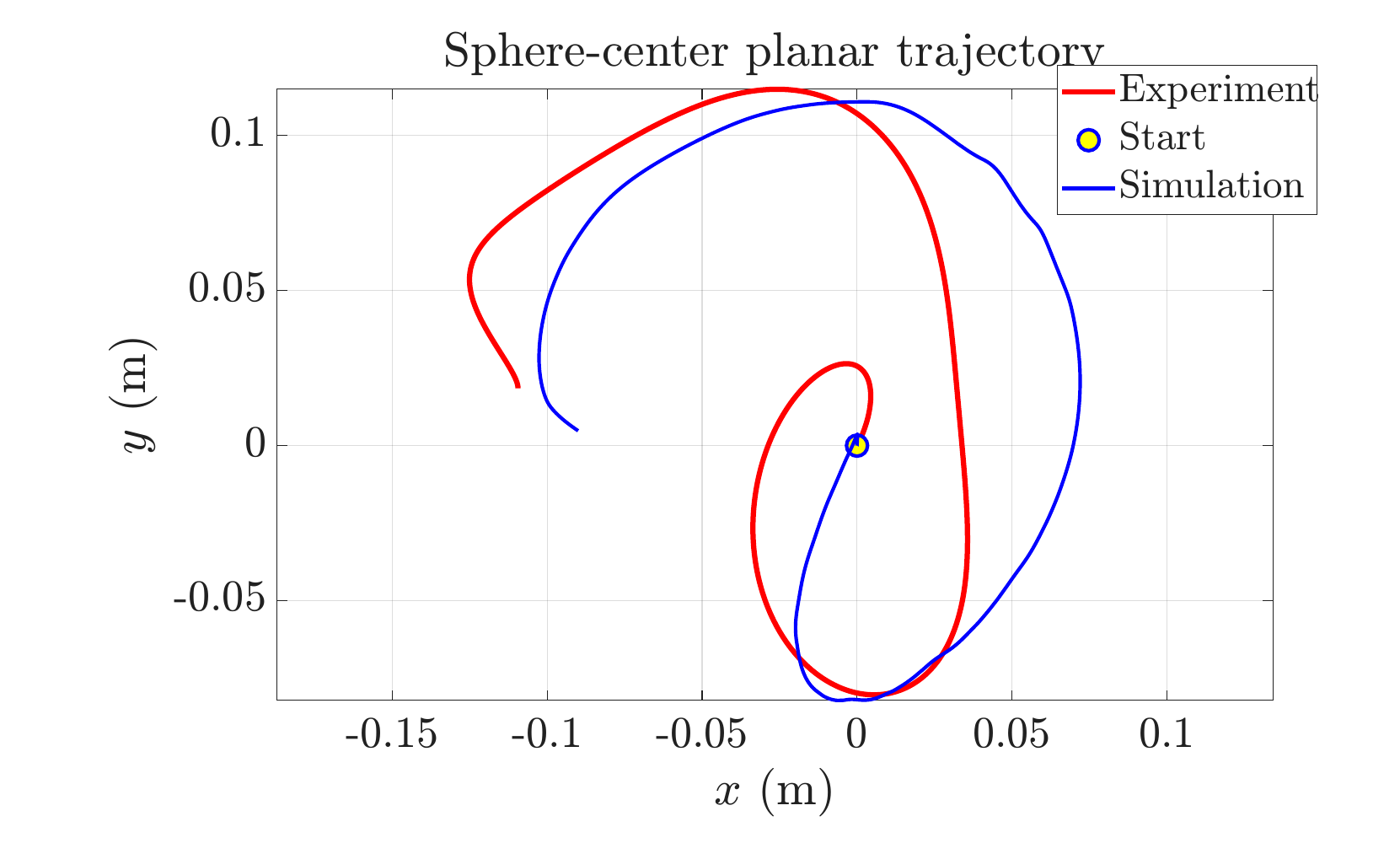}
    \caption{Experiment--simulation comparison of the sphere-center planar trajectory.}
    \label{fig:exp_sim_planar}
\end{figure}
\begin{figure*}[t!]
    \centering
      \includegraphics[width=2.6in]{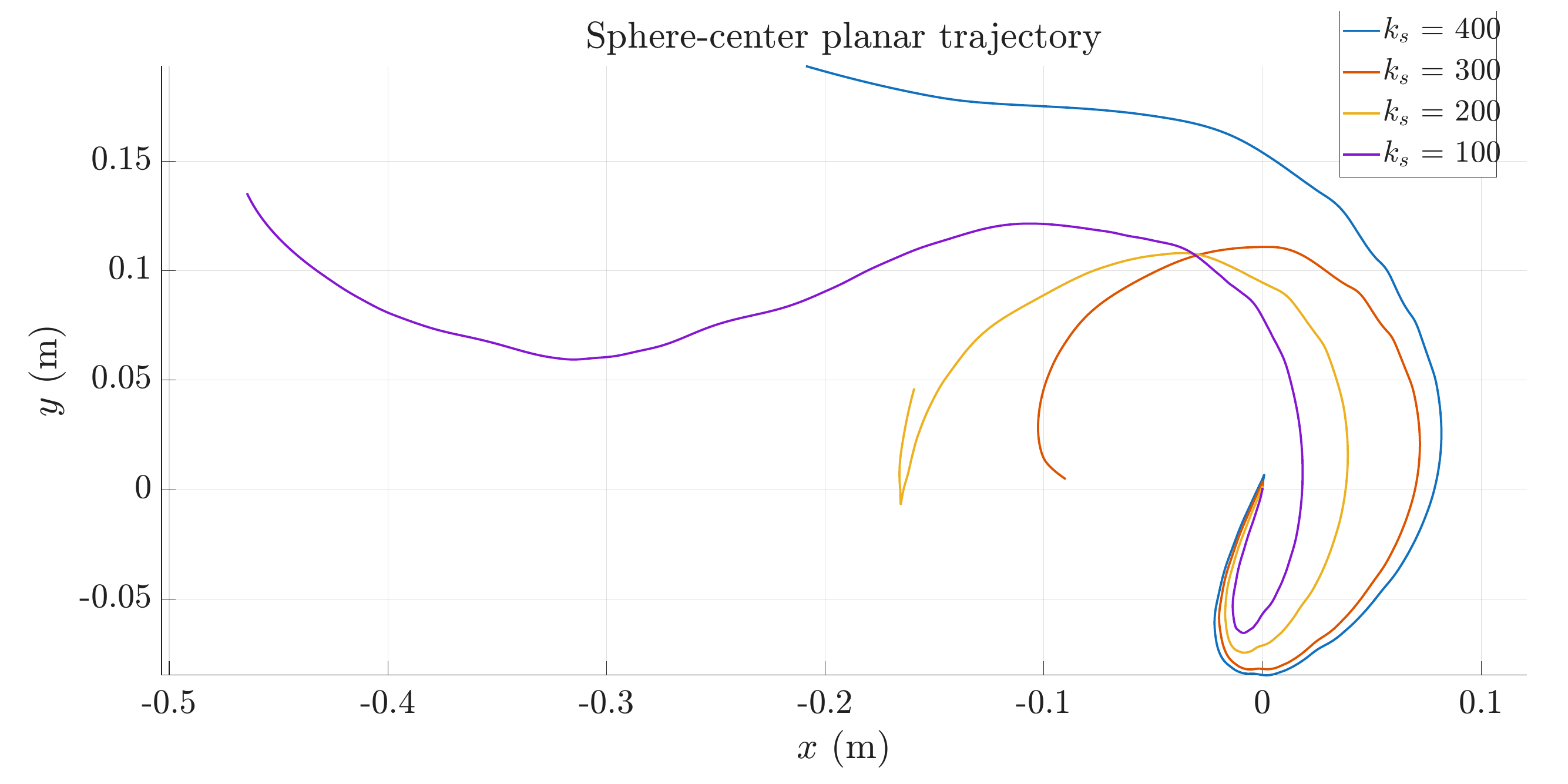}
     \includegraphics[width=2.1in]{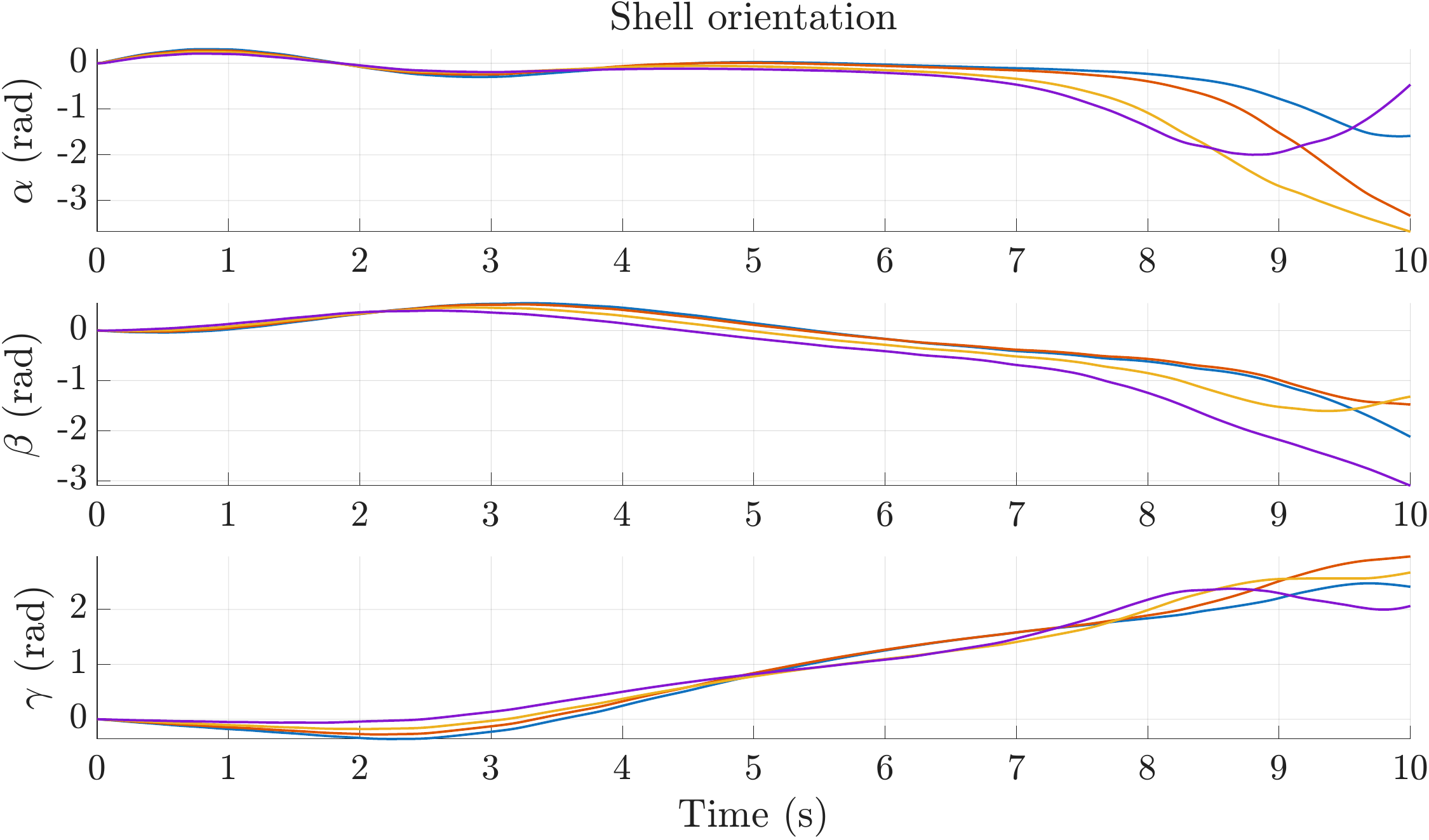}
    \includegraphics[width=2.1in]{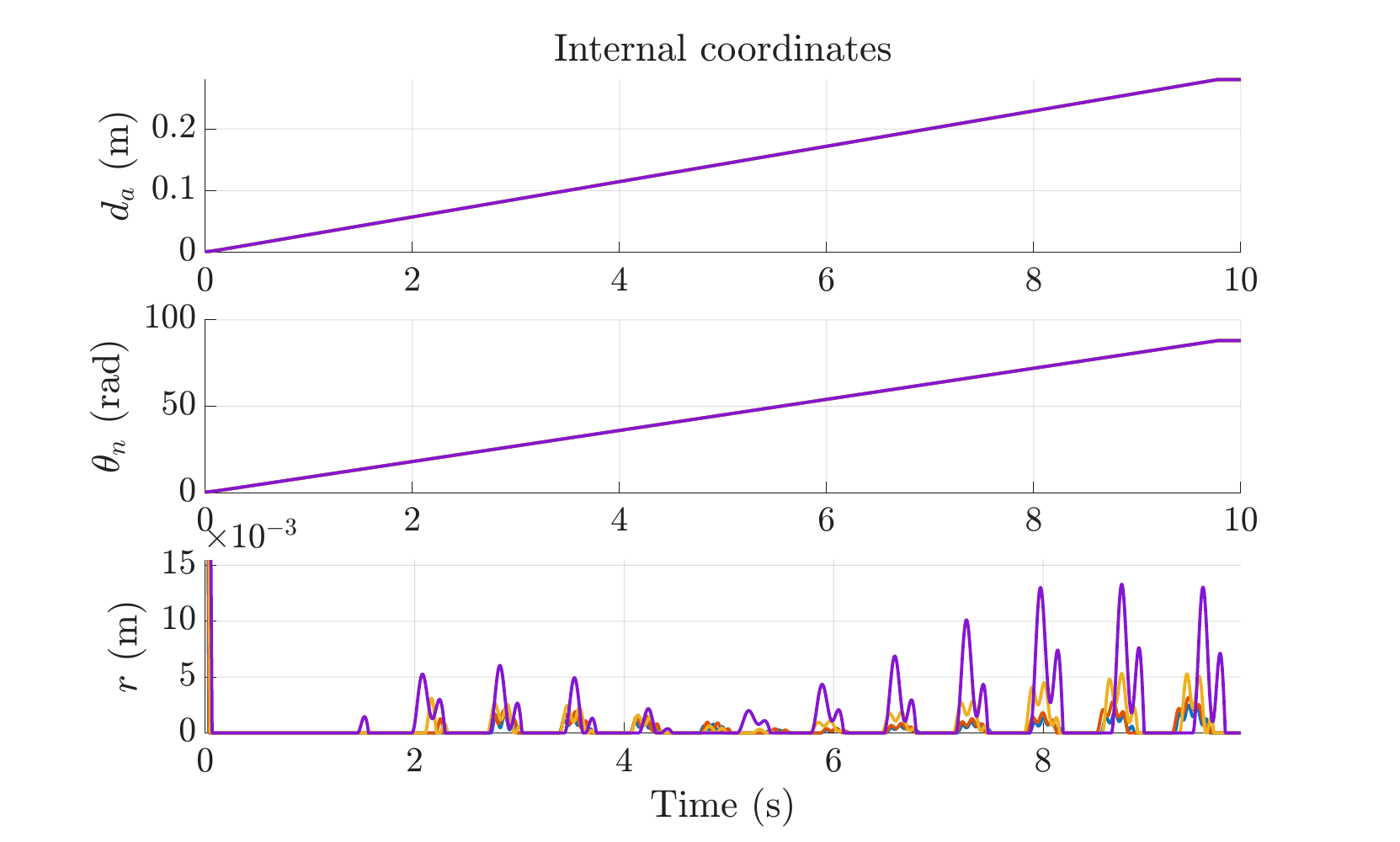}
\caption{Effect of spring stiffness \(k_s\) on the simulated MonoRollBot response for fixed moving mass. Left to right: planar trajectory, shell orientation, and internal coordinates. Changes in \(k_s\) mainly affect the radial motion \(r\), leading to different rolling paths and attitude responses.}
    \label{fig:stiffnessvarioation}
\end{figure*}

For the conservative configuration vector is obtained from the potential energy we obtain it as
\begin{align*}
\uvec{g}_q(\uvec{q})=\frac{\partial V}{\partial \uvec{q}}
\end{align*}
\begin{equation}={\small
\begin{bmatrix}
0\\[2pt]
0\\[2pt]
g m_c\bigl(-\Delta\sin\alpha\cos\beta + r\cos\beta\cos\alpha\sin\theta_n\bigr)\\[4pt]
g m_c\bigl(-\Delta\cos\alpha\sin\beta - r\cos\beta\cos\theta_n
-r\sin\beta\sin\alpha\sin\theta_n\bigr)\\[4pt]
0\\[4pt]
g m_c\cos\alpha\cos\beta\\[4pt]
g m_c r\bigl(\sin\beta\sin\theta_n + \cos\beta\sin\alpha\cos\theta_n\bigr)\\[4pt]
k_s r + g m_c\bigl(-\sin\beta\cos\theta_n + \cos\beta\sin\alpha\sin\theta_n\bigr)
\end{bmatrix}}
\label{eq:gqexplicit}
\end{equation}
where \(g\) is the gravitational acceleration, \(m_c\) is the lumped internal moving mass, \(k_s\) is the effective radial spring stiffness.  Finally, the generalized damping vector is obtained from the Rayleigh dissipation function as
\begin{equation}{\small
\uvec{d}_q(\dot{\uvec{q}})
=
\frac{\partial \mathcal{D}}{\partial \dot{\uvec{q}}}
=
\begin{bmatrix}
0&
0&
c_s\dot{\alpha}&
c_s\dot{\beta}&
c_s\dot{\gamma}&
c_d\dot{d}_a&
c_\theta\dot{\theta}_n&
c_r\dot{r}
\end{bmatrix}^T
\label{eq:dq}}
\end{equation}
where \(c_s\), \(c_d\), \(c_\theta\), and \(c_r\) are the viscous damping coefficients associated with shell attitude, screw translation, nut rotation, and radial mass motion, respectively.

On numerical solution based on mechanical properties, the constrained dynamics are integrated numerically by solving the augmented linear system \eqref{eq:augmented_system} at each evaluation of the right-hand side. Consistent initial conditions must satisfy
\begin{equation}
\uvec{A}(\uvec{q}_0)\dot{\uvec{q}}_0=\uvec{0},
\qquad
\phi_h(\uvec{q}_0)=0.
\label{eq:init_cons}
\end{equation}
In the implementation, this is enforced by choosing \(\dot{\uvec{q}}_0=\uvec{0}\) and setting \(d_{a,0}=a\theta_{n,0}\). Also , to represent the finite internal travel, the simulation also imposes the admissibility bounds
\begin{equation}
0 \le d_a \le 2z_{\mathrm{off}},
\qquad
0 \le \theta_n \le \frac{2z_{\mathrm{off}}}{a},
\label{eq:bounds1}
\end{equation}
\begin{equation}
0 \le r \le \sqrt{z_{\mathrm{off}}^2-\left(d_a-z_{\mathrm{off}}\right)^2}.
\label{eq:bounds2}
\end{equation}
These are not additional ideal constraints in the symbolic derivation; they are unilateral geometric bounds used in the numerical implementation to keep the internal mass inside the shell clearance. Geometric and mass parameters are taken from the prototype dimensions and component properties summarized in Table~\ref{tab:params}, while the damping terms are treated as effective lumped coefficients for the present study.

\begin{figure}[t!]
    \centering
      \includegraphics[width=3.3 in]{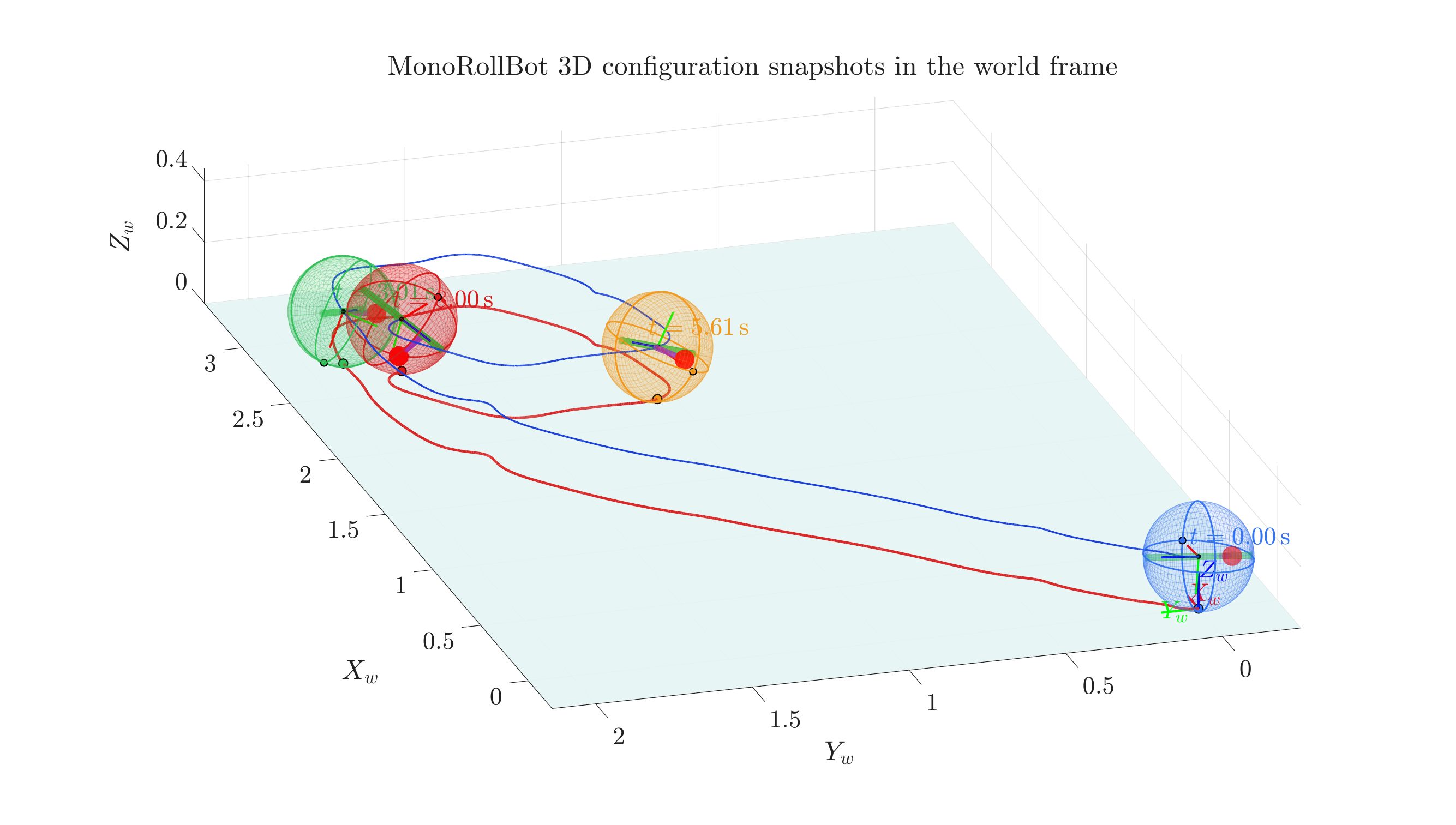}
      %
      %
      %
     \includegraphics[width=3.2 in]{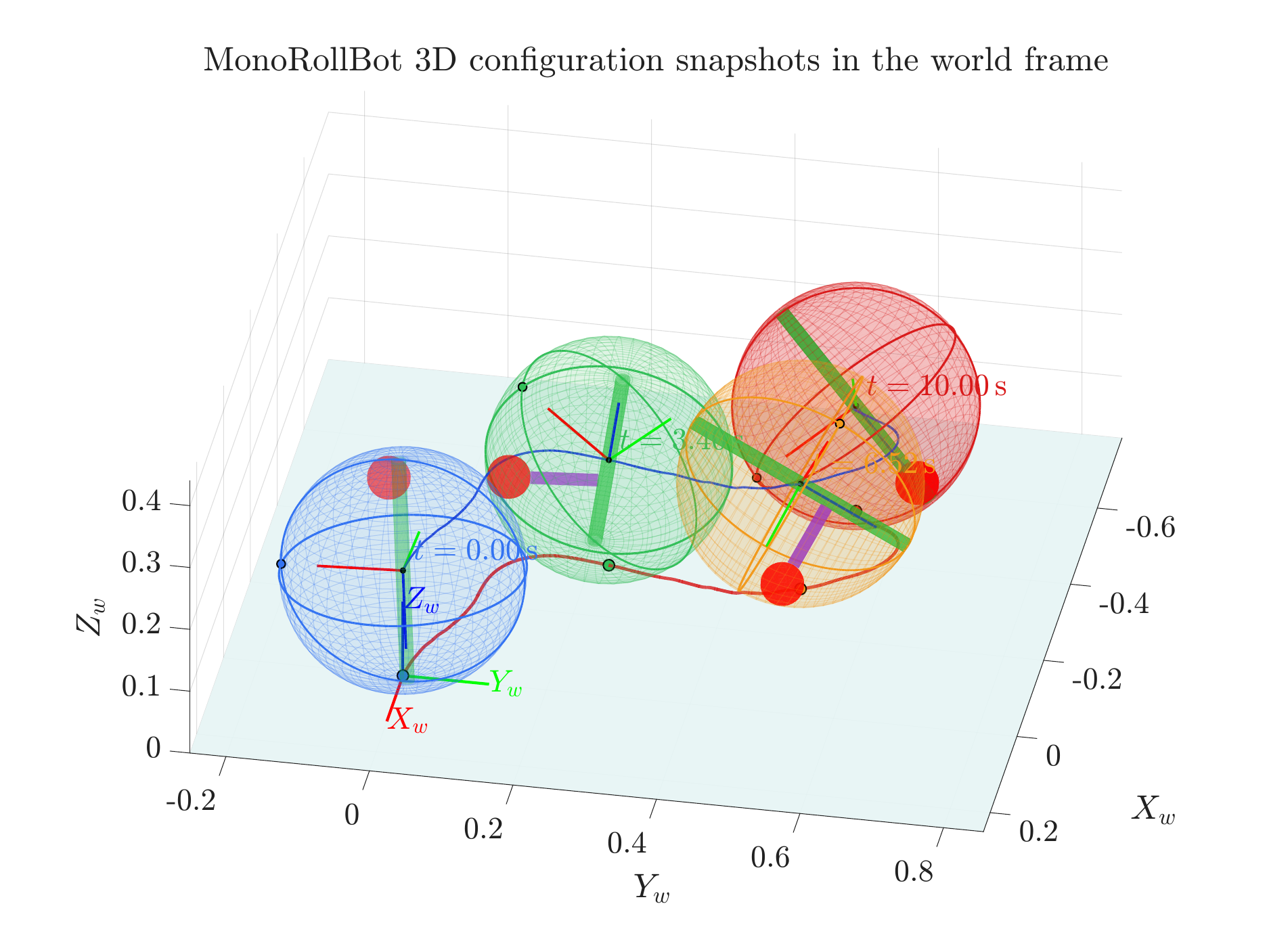}
\caption{Representative rolling motions obtained from two different initial configurations \(\uvec{q}_0\). The snapshots show that changing the initial internal mass phase and shell orientation can produce distinct locomotion patterns, including near-linear motion with gradual turning and wave-like rolling trajectories.}
    \label{fig:differentconfigurations}
\end{figure}
\section{Results and Discussion}
This section first evaluates the proposed full nonlinear constrained model through a representative simulation--experiment comparison and then uses the validated model for motion analysis of key design variables. The focus is on whether the formulation reproduces the dominant rolling trends of the prototype and on how compliance, screw-driven internal motion, and initialization affect the resulting locomotion patterns.

All simulations were performed using the augmented DAE model in \eqref{eq:augmented_system} and integrated numerically in MATLAB with \texttt{ode15s}. The computations were carried out on a 64-bit Windows workstation equipped with an Intel Core Ultra 9 275HX processor and 64\,GB RAM. Unless otherwise stated, the initial condition was set to \(\uvec{q}_0=\begin{bmatrix}0 & 0 & 5^\circ & 3^\circ & -\pi/2 & a(0.30) & 0.30 & 0.05\end{bmatrix}^{\T}\) with \(\dot{\uvec{q}}_0=\uvec{0}\), which satisfies both the screw consistency relation and the rolling velocity constraints at \(t=0\). This scenario corresponds to a motion in which the internal mass starts near the lower region of the sphere and progressively travels toward the opposite end through a full rotational cycle. In addition to the physical parameters listed in Table~\ref{tab:params}, the simulations respect the admissibility bounds in \eqref{eq:bounds1}--\eqref{eq:bounds2}, which keep the internal mass within the shell clearance without introducing additional ideal constraints into the dynamic model.

A representative validation case is first used to compare simulated and experimental rolling motion in the plane, since the sphere-center trajectory is the most directly observable output of the platform. The corresponding simulated state evolution is shown in Fig.~\ref{fig:fullstate_validation}, while Fig.~\ref{fig:sim_snapshots} illustrates the underlying mechanism in both \(\Sigma_w\) and \(\Sigma_b\): the lead screw produces axial repositioning of the internal mass, and the combined nut rotation and compliant radial motion generate a helical internal path that drives shell rolling. The constraint residuals remain small throughout the simulation, confirming accurate numerical enforcement of both the rolling and screw constraints. As shown in Fig.~\ref{fig:exp_sim_planar}, the model also reproduces the dominant measured planar trend, including the loop-like trajectory shape, turning direction, and characteristic motion scale under the same test conditions. Taken together, these results indicate that the full constrained model captures the main robot dynamics with sufficient fidelity for the motion studies that follow. As shown in Fig.~\ref{fig:fullstate_validation}, the constrained formulation is numerically well behaved: \(d_a\) and \(\theta_n\) remain consistent with the screw relation, and the residuals stay small, with rolling errors on the order of \(10^{-7}\) and the screw error on the order of \(10^{-15}\). Together with the planar agreement in Fig.~\ref{fig:exp_sim_planar}, this indicates that the full constrained model is sufficiently accurate for the parametric motion studies that follow.

Fig.~\ref{fig:differentconfigurations} further shows that MonoRollBot can generate qualitatively different rolling patterns simply by changing its initial internal configuration under the same actuation principle. In the first case, the robot is initialized at \(\uvec{q}_0=\begin{bmatrix}0 & 0 & 5^\circ-\pi & 3^\circ & 0 & a(0.30+7\pi) & 0.30+7\pi & 0.05\end{bmatrix}^{\T}\), corresponding to a configuration in which the lead screw is initially approximately horizontal with respect to the ground. This produces an initially near-linear rolling segment, similar to the dominant forward-tipping behavior often associated with pendulum-driven spherical robots \cite{tafrishi2025survey}. As the internal mass continues to evolve along the screw and around the shell, however, the rolling direction gradually bends and the path develops a spiral-like turning component. This indicates that MonoRollBot cannot be interpreted as a purely pendulum-driven system, since the coupled axial, azimuthal, and compliant radial motion continuously changes both the gravity direction and the inertia distribution as the mass moves toward more lateral and upper regions of the shell. Sustained forward locomotion in this mode would therefore benefit from a resetting or re-centering strategy for the internal mass. A different behavior is obtained for \(\uvec{q}_0=\begin{bmatrix}0 & 0 & 5^\circ-\pi & 3^\circ & -\pi/2 & a(0.30) & 0.30 & 0.05\end{bmatrix}^{\T}\), for which the core starts near the upper region of the sphere and the robot exhibits a wave-like rolling path rather than a quasi-straight progression. The snapshots suggest that this behavior is generated by a different phase relation between shell motion and internal mass evolution, so that the robot alternates its rolling tendency instead of maintaining one dominant tipping direction. From a motion-planning perspective, these results suggest that MonoRollBot can realize different locomotion primitives, including near-linear propagation, turning, and wave-like progression \cite{OrioloVendittelli2005,SvininHosoe2008}, through internal-state initialization alone, without changing the hardware architecture. This highlights the strong role of configuration selection in future planning and control of screw-driven super-underactuated spherical robots like MonoRollBot.

\section{Conclusion}
This paper presented the full nonlinear nonholonomic constrained dynamics of MonoRollBot, to the best of our knowledge one of the first mono-actuated and super-underactuated spherical rolling robots to be modeled in this form. The proposed formulation retains the complete generalized coordinates of shell translation, shell attitude, screw travel, nut rotation, and radial mass motion, while treating rolling without slipping as a Pfaffian nonholonomic constraint and the screw relation as a holonomic constraint. The resulting model is expressed as a nonlinear differential-algebraic system in the generalized accelerations and constraint multipliers. The model was then used for representative validation and motion analysis, showing how gravity, compliance, internal mass redistribution, and initialization jointly shape the rolling behavior of MonoRollBot. These results establish a first full nonlinear dynamic basis for understanding and analyzing this class of mono-actuated spherical robots.

Future work will focus on hybrid controller design that exploits different internal configurations and motion primitives to achieve richer and more reliable locomotion, including omni-directional maneuvering, path-following, and configuration-reset strategies for sustained motion.


\bibliographystyle{ieeetr}
\bibliography{references}

\end{document}